\documentclass[journal]{IEEEtran}
\ifCLASSINFOpdf
\else
\fi

\usepackage[pdftex]{graphicx}
\usepackage{amssymb}
\usepackage{amsmath}
\usepackage{subfigure}
\usepackage{multirow}
\usepackage{booktabs}
\usepackage{xcolor}

\newcommand{\rev}[1]{\textcolor{black}{#1}}

\begin{document}
%
\title{\rev{An Improved Anomaly Detection Model for Automated Inspection of Power Line Insulators}}
%
%
%

\author{Laya~Das,~
        Blazhe~Gjorgiev,~
        and~Giovanni~Sansavini~
\thanks{L. Das, B. Gjorgiev and G. Sansavini are with the Department of Mechanical and Process Engineering, ETH Z\"urich, 8092 Z\"urich, Switzerland
e-mail: sansavig@ethz.ch.}
}

%
%

\markboth{Journal of \LaTeX\ Class Files,~Vol.~14, No.~8, August~2015}%
{Shell \MakeLowercase{\textit{et al.}}: Bare Demo of IEEEtran.cls for IEEE Journals}
%



\maketitle

\begin{abstract}
Inspection of insulators is important to ensure reliable operation of the power system. Deep learning is being increasingly exploited to automate the inspection process by leveraging object detection models to analyse aerial images captured by drones. A purely object detection-based approach, however, suffers from class imbalance-induced poor performance, which can be accentuated for infrequent and hard-to-detect incipient faults. This article proposes the use of anomaly detection along with object detection in a two-stage approach for incipient fault detection in a data-efficient manner. An explainable convolutional one-class classifier is adopted for anomaly detection. The one-class formulation reduces the reliance on plentifully available images of faulty insulators, while the explainability of the model is expected to promote adoption by the industry. \rev{A modified loss function is developed that addresses computational and interpretability issues with the existing model, also allowing for the integration of other losses. The superiority of the novel loss function is demonstrated with MVTec-AD dataset.} The models are trained for insulator inspection with two datasets -- representing data-abundant and data-scarce scenarios -- in unsupervised and semi-supervised settings. The results suggest that including as few as five real anomalies in the training dataset significantly improves the model's performance and enables reliable detection of rarely occurring incipient faults in insulators. 
\end{abstract}

\def\abstractname{Note to Practitioners}
\begin{abstract}
This paper is motivated by the problems of class imbalance and the small size of fault patterns in employing deep learning for fault detection of insulators. In practical settings, the number of faulty insulators is much smaller than that of healthy insulators, resulting in class imbalance in the learning problem. In such a scenario, deep neural networks perform poorly for the under-represented classes, i.e., faulty insulators. The performance can be worse when the patterns characterising faults are localised to a small portion of the image, e.g., patches of flash-over on insulator disks, making it a difficult learning problem. In this article, the problem of insulator fault detection is separated into two stages, wherein the first stage detects all disks in insulators and the second stage performs classification of healthy and faulty disks. The article focusses on the second stage and adopts a one-class classification formulation, learning predominantly from healthy insulators. The proposed approach allows the detection of incipient faults with very few samples of faulty insulators, facilitating the deployment of the technology with reduced investment in data collection and curation. The anomaly detection model also highlights portions of the disk that it considers to be characteristic of faults. This explainable nature of the model addresses potential skepticism about deploying deep neural networks in the industry. The current model has a few false predictions and spurious explanations, which will be addressed in future work.
\end{abstract}


%
\IEEEpeerreviewmaketitle

\section{Introduction}
%
%
%
%

\IEEEPARstart{P}{ower} transmission insulators play a key role in ensuring safe operation of power systems by providing support to transmission lines and preventing leakage through the towers. Power system operators conduct regular inspections of insulators, typically through on-site observations for manual detection of faulty insulators. The rapid progress recently witnessed in deep learning, coupled with the availability of high-quality images acquired by unmanned aerial vehicles is enabling the automation of the inspection process \cite{liu2023insulator}. 

The authors in~\cite{liu2018insulator} adopted an object detection-based approach and trained a Faster Region proposal-based Convolutional Neural Network (Faster RCNN) model to detect three different types of insulators. In addition, their model detects missing caps of insulators and achieves a fault detection precision of $92\%$. Similarly, \cite{jiang2019insulator} trained a Single-Shot multi-box Detector (SSD) model to detect healthy and faulty insulators and obtained a precision of $92.48\%$. A densely connected feature pyramid network was used to modify the You Only Look Once version 3 (YOLOv3) model and improve detection accuracy in~\cite{zhang2021insudet}. The authors in~\cite{liu2021box} adopted a novel box-and-points-based representation of missing caps to train their model. A segmentation approach was coupled with Faster RCNN to identify missing caps in~\cite{zhao2021insulator}, which resulted in better performance. A particle swarm optimisation-based neural network partition algorithm was used to meet the real-time requirements of the application, resulting in $94.5\%$ accuracy and $58.5$ frames/second in~\cite{deng2022research}. A cross-stage partial and residual split attention network-based backbone was used along with a multiscale bidirectional feature pyramid network to modify the YOLOv4 model and achieve better detection performance in~\cite{hao2022insulator}. The authors in \cite{shakiba2022transfer} used a VGG-19 model to perform fault detection with CPLID and demonstrated better performance than competing YOLO models. A Gaussian prior was applied to the object detection heads in \cite{dai2022uncertainty}. The uncertainty scores provided by the model were used to refine the bounding box predictions and achieve better performance. The authors in \cite{zhou2023fault} used the attention mechanism, hourglass network and a rotation mechanism to improve the detection performance of the YOLOv5 model.

\rev{\subsection{Knowledge gaps} \label{sec:gaps}
Despite this increasing research attention, we identify key shortcomings in the state-of-the-art that have left significant knowledge gaps in the literature.
\paragraph{Small and less diverse datasets} Many object detection models are trained and evaluated with small datasets (e.g., $80$ images for training and $40$ images for testing \cite{liu2018insulator}, $581$ images for training and $388$ images for testing \cite{liu2021box}) with little or no investigation of the sensitivity to image features in the background and foreground. Thus, their applicability under more general settings of diverse features is not known. Further, a recent study reported that including several data augmentation techniques does not achieve a performance improvement \cite{das2022object}.
\paragraph{Class imbalance} Several articles suffer from class imbalance in their datasets (e.g., $600$ healthy insulators and $248$ faulty insulators in~\cite{zhang2021insudet,hao2022insulator}), which hampers the performance for under-represented classes, i.e., faulty insulators. This poses a challenge to automating the inspection process in a well-functioning grid. Since most insulators are expected to be healthy, collecting more images invariably leads to a class-imbalanced dataset. While the imbalance can be treated by undersampling the healthy insulators, this will prevent a major portion of the dataset from being used for training. Thus, addressing class imbalance in a data-efficient manner is needed, which requires a methodological improvement over a purely object detection-based strategy.
\paragraph{Simple detection tasks} The literature addresses the relatively simpler task of detecting missing caps/disks, ignoring incipient faults such as flashed and broken disks. The flash-over caused by electrical flashes or lightning strikes as well as irregularly shaped broken disks are important for the operator's situational awareness and informed prioritisation of maintenance activities. Further, these faults can be difficult to detect compared to missing caps, since these faults are localised to a relatively smaller portion in the image. The small patterns coupled with class imbalance further lead to poor performance of object detection models \cite{das2022object}.}

In this article, we address the above gaps with a two-stage approach that relies on detection and segmentation for automated inspection of insulators.


\subsection{Contributions}
This article makes the following contributions:
\begin{enumerate}
    \item \textit{Problem formulation:} We pose the detection of incipient faults, specifically flashed and broken disks as an anomaly detection problem. This methodological choice splits the detection problem into two stages and uses simplified models in each stage. The first model processes the aerial image to locate insulators and disks, while the second model processes images of disks to detect anomalous disks. This setting also drastically increases the number of images available for training the anomaly detection model.
    \rev{\item \textit{Improved anomaly detection:} We adopt a one-class classification-based anomaly detection strategy, and employ the Fully Convolutional Data Description (FCDD) model. We identify mathematical and computational challenges associated with the FCDD loss and introduce a novel loss function to tackle these challenges. The resulting modified FCDD provides a clear definition of class probabilities at the pixel level for each image, offering better interpretability and integration with complex loss functions. We demonstrate the superiority of modified FCDD over the original model with a standard anomaly detection dataset (MVTec-AD).
    \item \textit{Insulator inspection with anomaly detection:} We integrate the modified FCDD with focal loss to construct a compound loss function that adapts to the nuances of insulator inspection datasets, and compare the performance with original FCDD for two datasets representing data-abundant and data-scarce scenarios. We also show two methods of improving the performance of anomaly detection in the data-scarce scenario.}
\end{enumerate}
The rest of the article is organized as follows: Section \ref{sec:method} presents the methodology adopted in this work. Section \ref{sec:insualtor} presents the application of the anomaly detection model to insulator fault detection along with the dataset collection and curation procedure. The experimental setup is also discussed. Section~\ref{sec:results} presents the results of our experiments. Finally, Section~\ref{sec:conclusions} provides concluding remarks.

\section{\rev{Problem Formulation}} \label{sec:formulation}
\rev{In this section, we present our motivation to adopt an anomaly detection-based problem formulation to solve the fault detection problem for insulators and discuss how anomaly detection coupled with object detection can address the research gaps identified in Section \ref{sec:gaps}.}

\subsection{\rev{Insulator inspection with object detection}} \label{sec:motivation}
We address the problem of insulator fault detection, i.e., identifying the presence of faults in insulators. Specifically, we consider two types of faults -- flashed disks and broken disks, which are relatively hard-to-detect incipient faults and have not been extensively studied in the literature. A recent study showed that this problem can be elusive with an object detection-based approach in data-scarce scenarios~\cite{das2022object}. State-of-the-art object detection models were reported to perform very well for detecting insulators (which have high frequency and large size), relatively poorer for healthy disks (very high frequency but small size), and very poor for flashed disks (very low frequency and small size) \cite{das2022object}. \rev{Thus, class imbalance and scale imbalance are key performance-limiting factors for object detection models. As discussed in Section \ref{sec:gaps}, class imbalance challenge cannot be addressed by collecting more images. Scale imbalance is inherent to the application and cannot be avoided because each insulator is constructed with roughly $15$ disks, leading to an inevitable $15\times$ difference in scale}. This motivates the need for a methodologically novel approach to insulator fault detection.

A closer inspection of the performance of object detection models for the faulty disks reveals that the models exhibit high classification errors with rather small localisation errors. In other words, the models accurately localise the disks but encounter difficulty in correctly classifying them as healthy and faulty. \rev{Since object detection models learn the two tasks simultaneously, which can be difficult for imbalanced datasets, we investigate if segregating the localisation and classification tasks can benefit the fault detection objective.} 

\subsection{\rev{Insulator inspection and anomaly detection}}
\rev{
In a segregated approach, the object detection model learns to localise disks regardless of their state (healthy or faulty). A second model then processes the extracted disks and learns to classify them as healthy or faulty. The training of object detection models has been extensively studied in the literature. In this article, we focus on the second model for solving the classification problem. It must be noted that the second model processes only images of disks, and thus, does not face scale imbalance.} However, class imbalance between healthy and faulty disks still persists and needs to be carefully addressed.

\begin{figure*}
    \centering
    \includegraphics[width=0.8\textwidth]{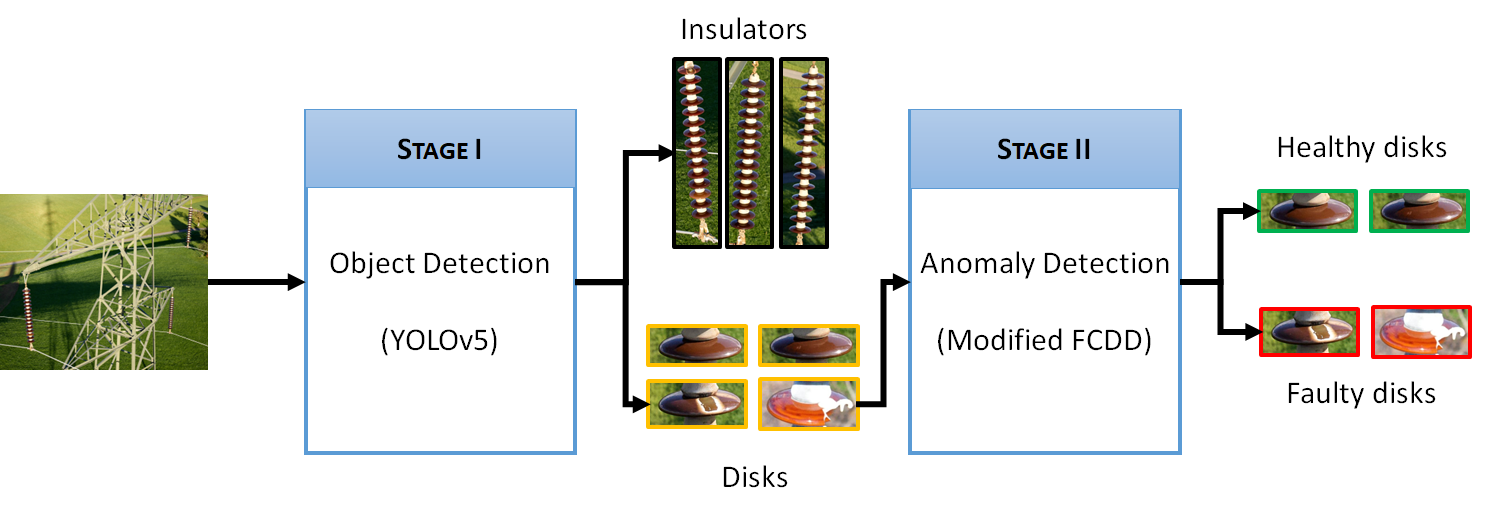}
    \caption{Two-stage insulator fault detection with object detection and anomaly detection. The texts in parentheses represent the model used in this work.}
    \label{fig:method}
\end{figure*}

The learning problem for the second model is to detect faulty disks, which occur very infrequently in a dataset with a large number of healthy disks, which naturally lends itself as an anomaly detection problem. Anomaly detection aims at detecting rarely occurring data points (anomalies) that deviate significantly from very frequently observed normal data points \cite{pang2021deep}. Anomaly detection is used in several applications such as blockchain networks \cite{hassan2022anomaly}, financial fraud detection \cite{hilal2022financial} and time series analysis \cite{schmidl2022anomaly}. This approach is also adopted for fault detection in industrial cyber-physical systems \cite{hao2021hybrid}, manufacturing \cite{maggipinto2022deep}, heavy-plate quality detection \cite{li2023self}, 
robotics \cite{schnell2022robigan}, electrical engineering \cite{chen2022fault}, and autonomous driving \cite{fang2023toward}.
This results in a two-stage approach that includes an object detection model followed by an anomaly detection model, as shown in Fig.~\ref{fig:method}.

\rev{
A two-stage problem formulation proposed here addresses the research gaps identified in Section \ref{sec:gaps} as follows:
\begin{enumerate}
    \item \textit{Small and less diverse datasets:} As reported in \cite{das2022object}, the classification errors of object detection models are the primary contributor to low performance in data-scarce scenarios. In a two-stage approach, the object detection model does not learn the classification (healthy and faulty) task. Thus, the performance is determined solely by localisation errors, which are reported to be low \cite{das2022object}.
    \item \textit{Class imbalance:} The formulation of the classification problem as anomaly detection explicitly addresses the problem of class imbalance.
    \item \textit{Simple detection tasks:} In this article, we consider the detection of hard-to-detect incipient faults, i.e., flashed and broken disks.
\end{enumerate}
}

\begin{figure}
    \centering
    \subfigure[Aerial images of insulators]{
        \includegraphics[width=0.2\textwidth,height=0.15\textwidth]{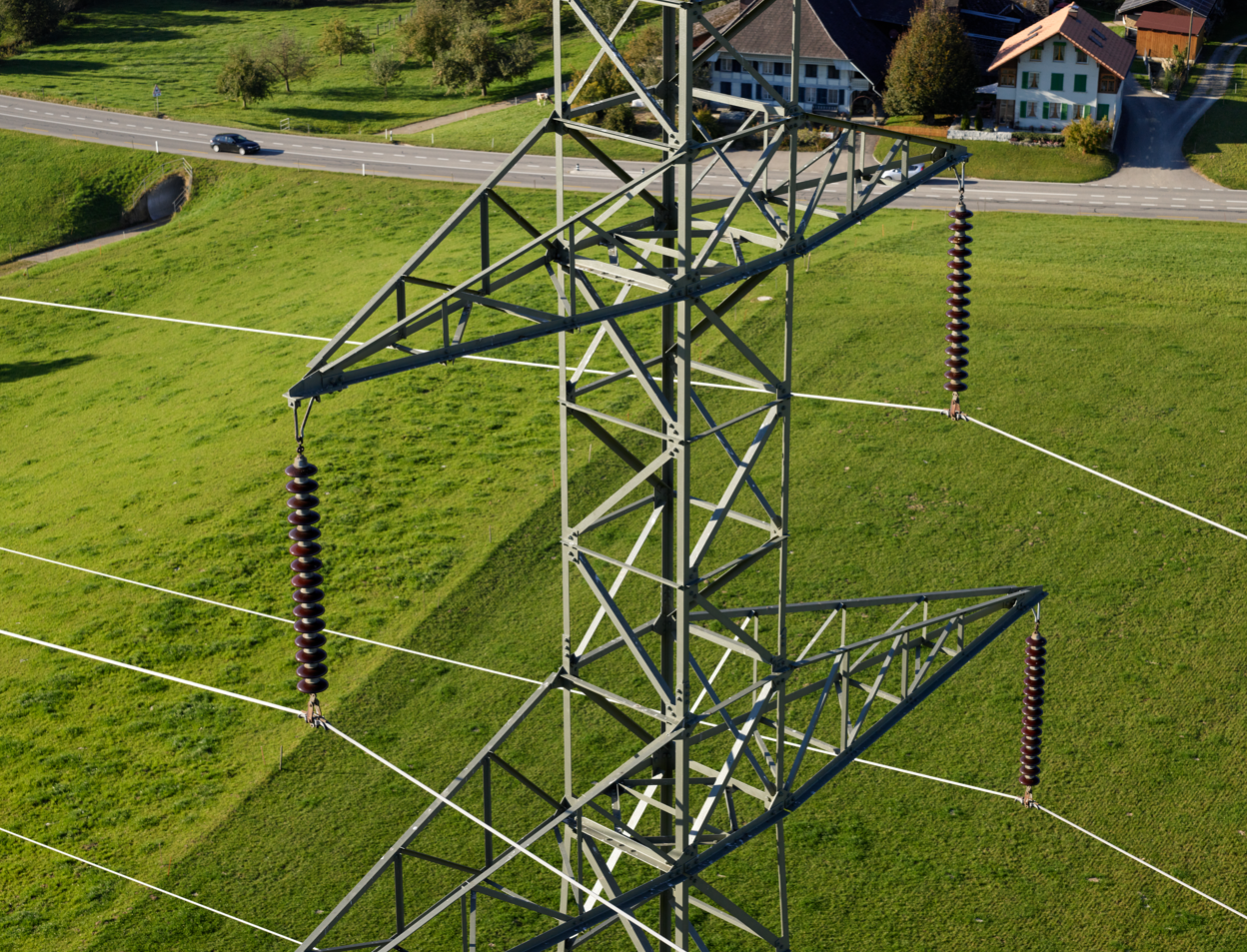}
        \includegraphics[width=0.2\textwidth,height=0.15\textwidth]{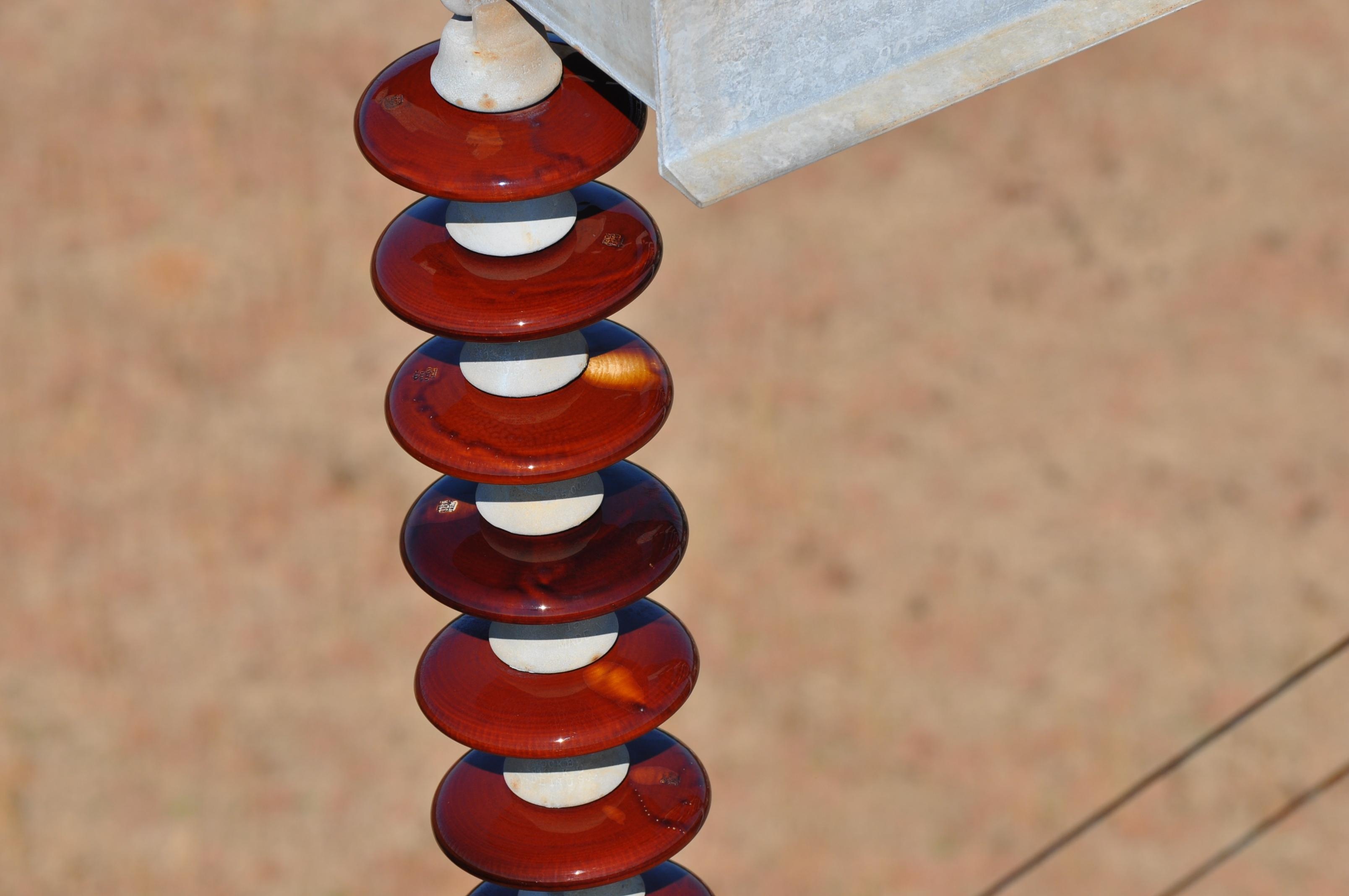}
    }
        \subfigure[Extracted images of disks]{
        \includegraphics[width=0.2\textwidth,height=0.1\textwidth]{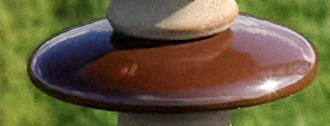}
        \includegraphics[width=0.2\textwidth,height=0.1\textwidth]{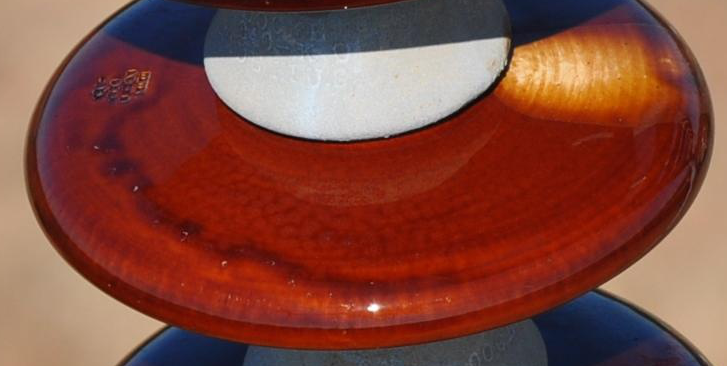}
    }
    \caption{Aerial images of insulators and extracted images of disks. The extracted images contain very little background compared to aerial images.}
    \label{fig:insulators_disks}
\end{figure}

\rev{
In addition, we note that the two-stage formulation has the following key impacts on the learning problem:
\begin{enumerate}
    \item The fault detection task is divided into simpler sub-tasks. In the first stage, a 2-class object detection model (to detect insulators and disks) will suffice, as opposed to a 4-class model (that detects insulators, healthy disks, flashed disks and broken disks) in a one-stage approach. This problem simplification makes it easier for the model to perform better. 
    \item In the second stage, the samples consist only of extracted images of accurately localised disks, which exhibit very little background compared to the first stage. This is evident from Fig.~\ref{fig:insulators_disks}, which shows sample aerial images of insulators and extracted images of disks. As a result, the anomaly detection model observes little interference from the background.
\end{enumerate}
}

\section{\rev{Methodology}}\label{sec:method}
In this section, we present the state-of-the-art one-class classifier adopted for building the anomaly detector, i.e., fully convolutional data description \cite{liznerski2021explainable}. We identify a few challenges with FCDD loss in semi-supervised training and propose a novel loss function that is aligned with cross-entropy to devise the modified FCDD.

Anomaly detection is a very well-studied problem, with several approaches available in the literature \cite{ruff2021unifying}. In this work, we are interested in anomaly detection for images of disks and thus adopt a computer vision-based approach to address this problem. A taxonomy and review of the algorithms for anomaly detection using deep learning is presented in \cite{pang2021deep}. Deep learning-based methods often adopt autoencoders for image \cite{zhou2017anomaly} and video \cite{zhao2017spatio} anomaly detection, which are trained on normal data points. Since such a model only sees normal data points during training, it is expected to exhibit high errors in reconstructing an anomalous data point, thereby allowing anomaly detection by monitoring the reconstruction error of the model. While such models are very popular, they do not offer a way to include any prior information about anomalies during training. However, in the case of detecting faulty disks, one has prior information about the flash-over patterns and shapes of broken disks that could be leveraged to improve the performance of the model. We therefore employ a state-of-the-art anomaly detection method that allows using the patterns observed in anomalous data points during training. The model also provides explanations for its predictions, which can promote its adoption by the industry.

\subsection{Anomaly detection with fully convolutional data description (FCDD)}
The FCDD is a state-of-the-art anomaly detection approach \cite{liznerski2021explainable} that has been demonstrated to provide excellent performance on standard datasets, including ImageNet and MVTec-AD (a dataset for detecting defects in manufacturing \cite{bergmann2019mvtec})\rev{, outperforming several contemporary anomaly detection methods. It allows both unsupervised and semi-supervised training, incorporating information about anomalies at the time of training.} FCDD is a one-class classification model that performs anomaly detection by mapping all normal data points in the vicinity of a centre $\mathbf{c_0}$ in the output space, which results in the anomalies being mapped away from $\mathbf{c_0}$. FCDD is based on the hypersphere classifier (HSC), which employs the following loss function:
\begin{equation}
\begin{split}
    \mathcal{L}_{HSC}=&\frac{1}{N}\sum_{i=1}^N\Big[\left(1-y_i\right)\tilde{h}\left(f\left(x_i;\theta\right)-\mathbf{c_0}\right)
    \\ &-y_i\log\left(1-\exp\left(-\tilde{h}\left(f\left(x_i;\theta\right)-\mathbf{c_0}\right)\right)\right)\Big]
\end{split}\label{eq:hsc}
\end{equation}
where $x_i\in\mathbb{R}^{c\times h\times w}$ represents the $i^{th}$ input with $c$ channels, height $h$ and width $w$. $y_i\in\{1,0\}$ represents the $i^{th}$ target such that $y_i=1$ for anomalous data points, and $y_i=0$ for normal data points, $N$ represents the number of data points, $f(\cdot;\theta)$ represents the neural network parameterised in $\theta$, and $\tilde{h}(\cdot)$ represents the pseudo-Huber loss, defined as $\tilde{h}(x)=\sqrt{x^2+1}-1$. FCDD uses a fully convolutional architecture for the neural network $f(\cdot;\theta)$, which transforms input $x_i\in\mathbb{R}^{c\times h\times w}$ to feature $z_i\in\mathbb{R}^{u\times v}$. In addition, the centre of the hypersphere is set to the bias of the last layer of $f$, so that the FCDD loss can be expressed as:
\begin{equation}
\begin{split}
    \mathcal{L}_{FCDD}=&\frac{1}{N}\sum_{i=1}^N\Bigg[\left(1-y_i\right)\frac{1}{u\cdot v}||\tilde{h}\left(z_i\right)||_1
    \\ &-y_i\log\left(1-\exp\left(-\frac{1}{u\cdot v}||\tilde{h}\left(z_i\right)||_1\right)\right)\Bigg]
\end{split}\label{eq:fcdd}
\end{equation}
where the quantity $\tilde{h}(z_i)$ is the pseudo-Huber loss of the features. This loss is summed for all entries of $\tilde{h}(z_i)$, and then normalised with respect to the number of entries, which provides a normalised measure of deviation of the feature from the centre. The above objective minimises the deviation for normal data points ($y_i=0$), and maximises the deviation for anomalous data points ($y_i=1$). The pseudo-Huber loss of features of a trained FCDD model can be considered as a heatmap of the input, with anomalous regions exhibiting higher values and normal regions exhibiting lower values. In general, the dimension of the features is lower compared to that of the input of a convolutional network, i.e., here $u<h, v<w$. To map this low-resolution heatmap to the original input image, FCDD further performs a deterministic upsampling of $\tilde{h}(z_i)$ with a Gaussian kernel~\cite{liznerski2021explainable}.

In the above formulation, the model takes only the images and labels, i.e., normal or anomalous, and learns a mapping that highlights the anomalous regions of an image. When anomalous images are used for training in this setting, information about patterns observed in faulty images is implicitly provided to the model. It is also possible to explicitly incorporate prior knowledge at the time of training, which can be achieved by modifying the FCDD loss to allow semi-supervised training as follows \cite{liznerski2021explainable}:
\begin{equation}
\begin{split}
    \mathcal{L}_{FCDD}^{SS}=&\frac{1}{N}\sum_{i=1}^N\left[\frac{1}{w\cdot h}\sum_{j=1}^{w\cdot h}\left(1-y_{i,j}\right)\tilde{h}'(z_i)_j\right.
    \\ &\left. -\log\left(1-\exp\left(-\frac{1}{w\cdot h}\sum_{j=1}^{w\cdot h}y_{i,j}\tilde{h}'(z_i)_j\right)\right)\vphantom{\frac{1}{1\cdot h}}\right]
\end{split}\label{eq:fcdd_semi}
\end{equation}
where $y_i\in\{1,0\}^{h\times w}$ is the ground truth map for the $i^{th}$ image and contains $1$ for all pixels exhibiting an anomaly, and $\tilde{h}'(z_i)\in\mathbb{R}^{h\times w}$ represents the up-sampled heatmap. The above loss function has been shown to result in better performing models \cite{liznerski2021explainable}.

\rev{\subsection{Modified FCDD}}
\rev{
In the section, we first motivate the need for a modified FCDD formulation by highlighting the drawbacks of the semi-supervised formulation and then present a novel loss function to address these issues. We recall the expression for binary cross-entropy (BCE) loss:
\begin{equation}
    \mathcal{L}_{BCE}=-\frac{1}{N}\sum_{i=1}^Ny_i\log(p_i)+(1-y_i)\log(1-p_i) \label{eq:bce}
\end{equation}
We see in Eq.~\ref{eq:bce} that we have a clear definition of the predicted class probability of the input, i.e., $p$ and $1-p$ for the BCE loss. We note two differences between the BCE and FCDD losses. First, the class labels ($y$ and $1-y$) have been swapped in formulating the FCDD loss. Although this notational difference has no impact on training, we highlight this difference for clarity in the ensuing discussion. To highlight the second difference, we rewrite Eq.~\ref{eq:fcdd_semi} as:
\begin{equation}
\begin{split}
    \mathcal{L}_{FCDD}^{SS}=&-\frac{1}{N}\sum_{i=1}^N\left[\frac{1}{w\cdot h}\sum_{j=1}^{w\cdot h}\left(1-y_{i,j}\right)\log(\exp(-\tilde{h}'(z_i)_j)\right.
    \\ &\left. +\log\left(1-\exp\left(-\frac{1}{w\cdot h}\sum_{j=1}^{w\cdot h}y_{i,j}\tilde{h}'(z_i)_j\right)\right)\vphantom{\frac{1}{1\cdot h}}\right]
\end{split}\label{eq:fcdd_semi_exp}
\end{equation}
Comparing Eq. \ref{eq:fcdd_semi_exp} with Eq. \ref{eq:bce}, we see that in the first term, the pixel-level predicted probability for the normal class can be expressed as $p=\exp(-\tilde{h}'(z_i)_j)$. The predicted probability for the anomalous class should then be $1-p=1-\exp(-\tilde{h}'(z_i)_j)$. However, this is not reflected in the second term in Eq. \ref{eq:fcdd_semi_exp}. Further, by construction, the second term does not allow us to obtain a clear definition of the pixel-level predicted probability for the anomalous class (because the summation is inside the $\exp(\cdot)$ function). Thus, the two terms in Eq. \ref{eq:fcdd_semi_exp} do not agree on the predicted probabilities at the pixel level. This disagreement prevents us from employing advanced loss functions such as focal loss~\cite{lin2017focal} that, for example, address class imbalance by re-weighing the loss terms for the normal and anomalous pixels. We also observe that Eq.~\ref{eq:fcdd_semi_exp} leads to computational issues for normal images. Specifically, for an image with no anomalous pixels, i.e., when $y_{i,j}=0$ $\forall i,j$, the second term in Eq.~\ref{eq:fcdd_semi_exp} evaluates to $-\infty$, preventing gradient calculation and backpropagation.
}

\rev{
To address these challenges, we formulate the modified FCDD loss for semi-supervised training that is aligned with BCE loss as:
\begin{equation}
\begin{split}
    \bar{\mathcal{L}}_{FCDD}^{SS}=&\frac{1}{N}\sum_{i=1}^N-\left[\frac{1}{w\cdot h}\sum_{j=1}^{w\cdot h}\left(1-y_{i,j}\right)\log(\exp(-\tilde{h}'(z_i)_j)\right.
    \\ &\left. +y_{i,j}\log\left(1-\exp\left(-\tilde{h}'(z_i)_j\right)\right)\vphantom{\frac{1}{1\cdot h}}\right]
    \\ =&\frac{1}{N}\sum_{i=1}^N\frac{1}{w\cdot h}\sum_{j=1}^{w\cdot h}\left[\left(1-y_{i,j}\right)\tilde{h}'(z_i)_j\right.
    \\ & \left. -y_{i,j}\log\left(1-\exp\left(-\tilde{h}'(z_i)_j\right)\right)\vphantom{\frac{1}{1\cdot h}}\right]
\end{split}\label{eq:modified_fcdd}
\end{equation}
In Eq.~\ref{eq:modified_fcdd}, both terms have a consistent definition of the probability, $p_{i,j}=\exp(-\tilde{h}'(z_i)_j)$ at the pixel level. The modified loss also prevents any computational issues in gradient calculation. Note that the unsupervised FCDD loss in Eq.~\ref{eq:fcdd} is already aligned with BCE loss, and thus we do not make any changes to unsupervised training.}

\rev{With a consistent definition of the pixel-level probability $p_{i,j}=\exp(-\tilde{h}'(z_i)_j)$, we can now seamlessly integrate the modified FCDD with other advanced loss functions such as the focal loss:
\begin{equation}
\begin{split}
    \bar{\mathcal{L}}_{FCDD-FL}^{SS}=&\frac{1}{N}\sum_{i=1}^N\frac{1}{w\cdot h}\sum_{j=1}^{w\cdot h}\left[\left(1-y_{i,j}\right)\tilde{h}'(z_i)_j(1-p_{i,j})^\gamma\right.
    \\ & \left. -y_{i,j}p_{i,j}^\gamma\log\left(1-p_{i,j}\right)\vphantom{\frac{1}{1\cdot h}}\right]
\end{split}\label{eq:modified_fcdd_fl}
\end{equation}
where $\gamma\geq0$ represents the focussing parameter. This loss function allows penalising incorrect predictions to address an inevitable imbalance between the number of normal and anomalous pixels in a dataset. 
}

\section{Insulator Fault Detection with FCDD} \label{sec:insualtor}
In this work, we use FCDD in unsupervised and semi-supervised formulations (Eqs.~\eqref{eq:fcdd}~and~\eqref{eq:fcdd_semi}) \rev{along with the modified FCDD (Eqs. \eqref{eq:modified_fcdd} and \eqref{eq:modified_fcdd_fl})} to differentiate between healthy and faulty disks. In the unsupervised formulation, it is possible to learn from only normal samples, i.e., without including any anomalies at the time of training. It is also possible to use images of any object other than disks as anomalous samples, or employ an outlier exposure algorithm to synthetically generate anomalies, both of which improve the performance~\cite{liznerski2021explainable}. Finally, it is possible to include real anomalies (without any ground truth anomaly maps) to minimise $\mathcal{L}_{FCDD}$. In all these scenarios, either no \textit{real anomalies} are used during training, or the ground truth anomaly maps of real anomalies are not provided, resulting in unsupervised training. We perform unsupervised training with (i) no faulty disks and (ii) real faulty disks without anomaly maps. In addition, we also perform semi-supervised training by minimising $\mathcal{L}_{FCDD}^{SS}$, wherein we provide the ground truth anomaly masks during training. 

\subsection{Dataset preparation}
\rev{We use the MVTec-AD dataset for comparing the modified FCDD with the original version to establish the superiority of modified FCDD on a benchmark dataset. MVTec-AD contains $15$ types of real-life objects including capsules, screws, bottles, cables, etc. with and without anomalies, and is used as a standard dataset for anomaly detection. More details regarding this dataset can be found in \cite{bergmann2019mvtec}.}

\begin{figure}
    \centering
    \subfigure[Healthy disk and its map]{
        \includegraphics[width=0.2\textwidth,height=0.1\textwidth]{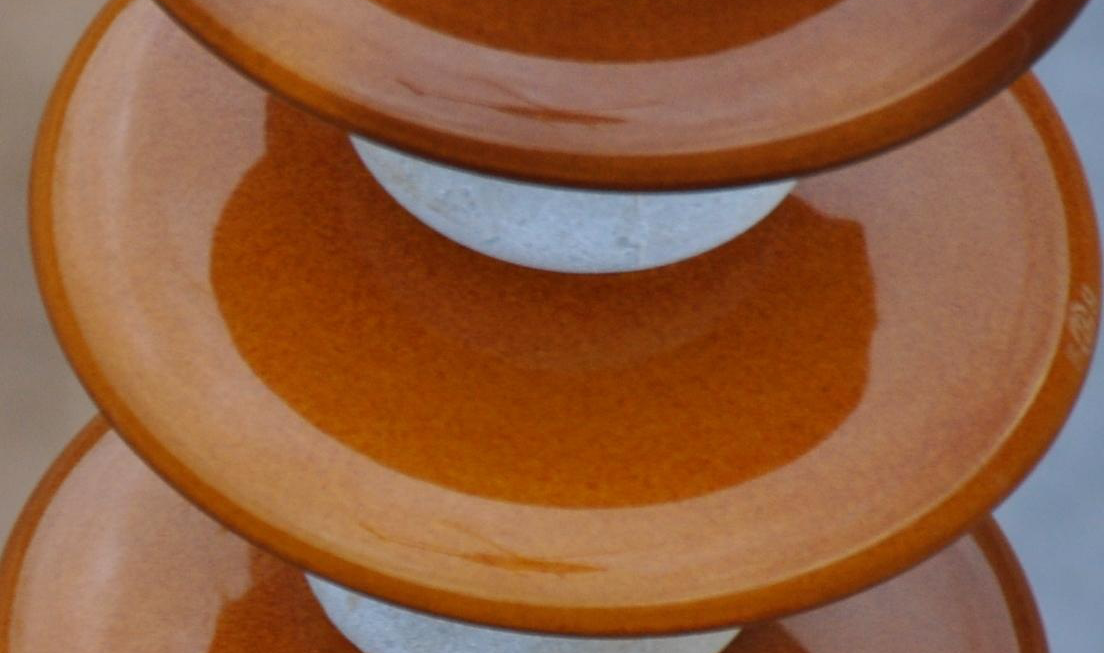}
        \includegraphics[width=0.2\textwidth,height=0.1\textwidth]{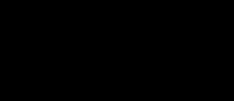}
    }
    \subfigure[Flashed disk and its map]{
        \includegraphics[width=0.2\textwidth,height=0.1\textwidth]{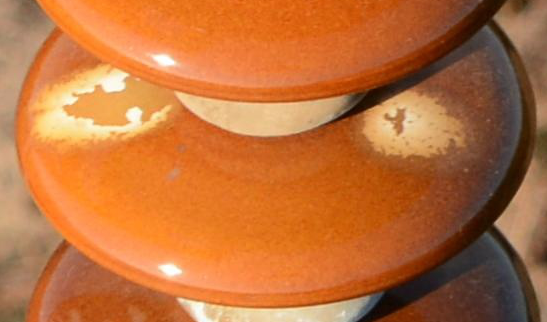}
        \includegraphics[width=0.2\textwidth,height=0.1\textwidth]{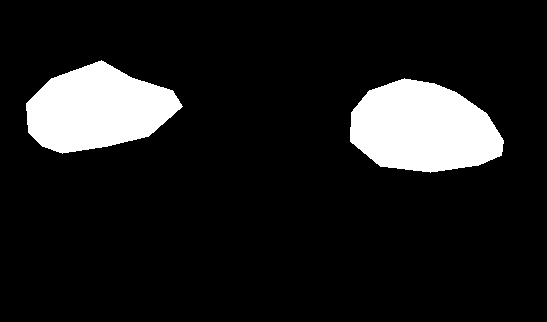}
    }
    \subfigure[Broken disk and its map]{
        \includegraphics[width=0.2\textwidth,height=0.1\textwidth]{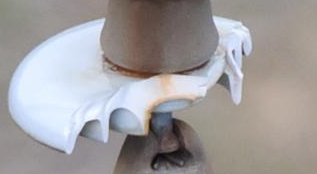}
        \includegraphics[width=0.2\textwidth,height=0.1\textwidth]{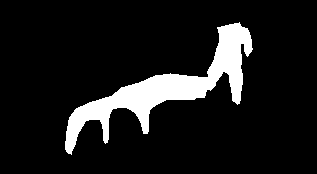}
    }
    \caption{Sample disks (healthy, flashed and broken), and their corresponding ground truth maps for semi-supervised training.}
    \label{fig:labels}
\end{figure}

We investigate the performance of FCDD for insulator inspection with two datasets. The first dataset is an openly available dataset curated by the Electric Power Research Institute and is referred to as the Insulator Defect Image Dataset (IDID) \cite{epriDS2021}. This dataset consists of $1600$ images of insulators with a variety of disk colours. Each image in this dataset has four orientations -- original, diagonally flipped, horizontally flipped and vertically flipped. In this work, we use only the images with original orientations, which results in $400$ images with $3286$ healthy disks, $716$ flashed disks and $282$ broken disks. This dataset is used to study the performance of anomaly detection in data-abundant scenario. The second dataset (referred to as SG) is a collection of $77$ aerial images collected by Swissgrid AG, with $2429$ healthy disks, only $53$ flashed disks and no broken disks. This dataset is used as the data-scarce scenario in this study. Since semi-supervised training requires the ground truth anomaly maps, these datasets are manually labelled with the ``labelme" tool \cite{Wada_Labelme_Image_Polygonal}. This involves generating polygonal segmentation masks for flash-over patches and regions of broken disks. The segmentation masks are then converted to $\{0,1\}^{h\times w}$ to obtain the ground truth map, which are shown in Fig.~\ref{fig:labels} for healthy, flashed and broken disks. Note that the maps for healthy disks are completely black and can be generated automatically. This restricts the manual burden of labelling to only faulty disks.

\subsection{Experimental setup}
\rev{We first study the performance of modified FCDD with MVTec-AD dataset (Section \ref{sec:results-mvtec_ad}). We use the default settings of the FCDD repository to generate the results only by modifying the loss function.} We then study the performance of the FCDD model on insulator datasets, trained individually on IDID and SG dataset, and compare the performance in data-abundant (Section \ref{sec:results-idid}) and data-scarce (Section \ref{sec:results-sg}) scenarios. \rev{We also compare the performance of unsupervised FCDD with an autoencoder-based anomaly detection model (AE).} Then, we investigate whether data from a different dataset can be used to improve the performance in a data-scarce scenario (Section \ref{sec:results-idid-sg}). To that end, we include data from both IDID and SG datasets to train a third FCDD model and examine if it performs better compared to the model trained only on the SG dataset. Note that while IDID has two types of faulty disks (flashed and broken), the SG dataset has only flashed disks. Thus, we train an FCDD model for the IDID dataset with both types of faulty disks for IDID, and use only flashed disks of IDID to augment the performance of (the third) FCDD for the SG dataset. The number of training and testing images used for different experiments are listed in Table~\ref{tab:data_split}. As an illustrative case, for unsupervised training without real anomalies for IDID, we use $2586$ healthy disks, while for semi-supervised training, we additionally use $200$ flashed disks and $100$ broken disks.

\begin{table}[]
    \centering
    \small
    \caption{Number of training and testing samples used in the three experiments. Unsupervised training uses all healthy samples along with 1) no anomalous samples or 2) anomalous samples without ground truth maps. Semi-supervised training uses all healthy and anomalous samples with the ground truth maps. ((F): Flashed disks, (B): Broken disks).}
    \begin{tabular}{ccccc}
        \toprule
          \multirow{2}{*}{Dataset} & \multicolumn{2}{c}{\#Healthy Samples} & \multicolumn{2}{c}{\#Anomalous Samples}\\
         \cline{2-5}
         & Train & Test & Train & Test \\
         \midrule
         SG & 2379 & 50 & 5 & 47 \\
         \multirow{2}{*}{IDID} & \multirow{2}{*}{2586} & \multirow{2}{*}{700} & 200 (F) & 516 (F) \\
         & & & + 100 (B) & + 182 (B) \\
         SG & 2379 & 50 & 5 & 47 \\
         + IDID & 0 & 0 & + 716 (F) & 0 \\
         \bottomrule
    \end{tabular}
    \label{tab:data_split}
\end{table}

All experiments are performed with the FCDD package \cite{liznerski2021explainable}. The default settings of FCDD have been employed, of which the notable settings are an input image size of $3\times224\times224$, the VGG-11-BN-based deep convolutional model~\cite{simonyan2014very}, a batch size of $128$, $200$ epochs for training and stochastic gradient descent optimizer. All experiments are performed on a Windows workstation with one NVIDIA RTX $3070$ GPU and $128$ GB of RAM. To evaluate and compare the performance of the models, we employ the standard area under the receiver operating characteristics curve (AUC) metric. The AUC can be expressed as:
\begin{equation}
    AUC=\int\limits_\mathcal{T} R_{TP}dR_{FP},\label{eq:auc}
\end{equation}
where $R_{TP}$ and $R_{FP}$ are the true positive rate and false positive rate, respectively, and are obtained as:
\begin{eqnarray}
    R_{TP}&=&\frac{\text{TP}}{\text{TP}+\text{FN}}\nonumber
    \\ R_{FP}&=&\frac{\text{FP}}{\text{FP}+\text{TN}}\nonumber
\end{eqnarray}
where TP, TN, FP and FN represent true positives, true negatives, false positives and false negatives, respectively. $\mathcal{T}$ represents the set of all decision thresholds used to predict the positives and negatives. First, different values of true and false positive rates are obtained for different classification thresholds, and AUC is then calculated according to Equation~\ref{eq:auc}. The optimal decision threshold can be obtained from the receiver operating characteristics by identifying the top-left point of the curve, which can then be used to evaluate the classification accuracy of the model. To account for uncertainty during training, FCDD trains $5$ different instances of a model. We report the average performance (AUC and optimal accuracy) of the five models in each experiment.

\section{Results} \label{sec:results}
In this section, we \rev{first compare the original and modified versions of FCDD for anomaly detection on the MVTec-AD dataset. We then present an evaluation and comparison of FCDD models trained for insulator inspection in} data-abundant and data-scarce scenarios with different formulations. Finally, we present the explanations provided by the semi-supervised FCDD model for its predictions.

\rev{\subsection{Anomaly detection on MVTec-AD} \label{sec:results-mvtec_ad}
In this section, we use the AUC calculated at the pixel level, referred to as GTMAP AUC (AUC for ground truth maps). We adopt this metric following the original results of FCDD reported in~\cite{liznerski2021explainable} for a fair comparison. The GTMAP AUC of original and modified versions of semi-supervised FCDD models are listed in Table~\ref{tab:mvtec}. The numbers for FCDD are taken from Table 2 of the original paper~\cite{liznerski2021explainable}. It can be seen that compared to FCDD, the modified FCDD is superior for $7$ objects, comparable for $6$ objects, and inferior for $2$ objects. This results in a net performance improvement of $7\%$ in total ($0.47\%$ on average) for the $15$ classes in the dataset. Furthermore, the worst-case performance for the dataset is improved by $2$ points, from $90\%$ for FCDD (Transistor) to $92\%$ for modified FCDD (Capsule). These results show that, besides being more aligned with BCE loss, the modified FCDD loss can lead to better model performance. In the following, we present the anomaly detection results for insulator inspection with the original FCDD and compare the performance with modified FCDD.
\begin{table}[]
    \centering
    \small
    \caption{\rev{GTMAP AUC of FCDD and modified FCDD with semi-supervised training for MVTec-AD dataset (\textbf{Bold} entries represent the best performance).}}
    {\color{black}\begin{tabular}{lcc}
        \toprule
         Object & FCDD & Modified FCDD \\
         \midrule
         Bottle & $0.96$ & $\mathbf{0.97}$ \\
         Cable & 0.93 & $\mathbf{0.95}$ \\
         Capsule & $\mathbf{0.95}$ & $0.92$ \\
         Carpet & $0.99$ & $0.99$ \\
         Grid & $0.95$ & $0.95$ \\
         Nut & $0.97$ & $\mathbf{0.98}$ \\
         Leather & $0.99$ & $0.99$ \\
         Metal Nut & $0.98$ & $0.98$ \\
         Pill & $0.97$ & $\mathbf{0.98}$ \\
         Screw & $0.93$ & $0.93$ \\
         Tile & $0.98$ & $\mathbf{0.99}$ \\
         Toothbrush & $\mathbf{0.95}$ & $0.94$ \\
         Transistor & $0.90$ & $\mathbf{0.93}$ \\
         Wood & $0.94$ & $\mathbf{0.96}$ \\
         Zipper & $0.98$ & $0.98$ \\
         \bottomrule
    \end{tabular}}
    \label{tab:mvtec}
\end{table}
}

\subsection{\rev{Insulator inspection with unsupervised learning}} \label{sec:results-idid}
In this section, we report the performance of models trained in an unsupervised manner, i.e., without ground truth for anomalous data points (original FCDD, Eq. \ref{eq:fcdd}).
\paragraph{\rev{Data abundance}} The AUC of FCDD without any faulty samples is $0.59$ for IDID, corresponding to a classification accuracy of $57.35\%$ for the optimal decision threshold. \rev{In comparison, an autoencoder-based anomaly detection model provides an AUC of $0.66$ and an optimal accuracy of $62.07\%$. While the AE model performs better than FCDD, it does not allow training with information from real anomalies.} Inclusion of only flashed disks as anomalous data points during 
training increases the AUC and optimal accuracy of FCDD to $0.79$ and $71.73\%$, respectively. This is a considerable performance improvement achieved with only $200$ training images of flashed disks and underscores the significance of using real anomalies during training. In addition to flashed disks, including broken disks during training yields an AUC of $0.75$, corresponding to an accuracy of $68.90\%$. This further demonstrates that FCDD can learn from multiple types of anomalies in the training dataset, and deliver performance comparable with learning from only one type of anomaly.

\begin{figure}
    \centering
    \includegraphics[width=0.5\textwidth]{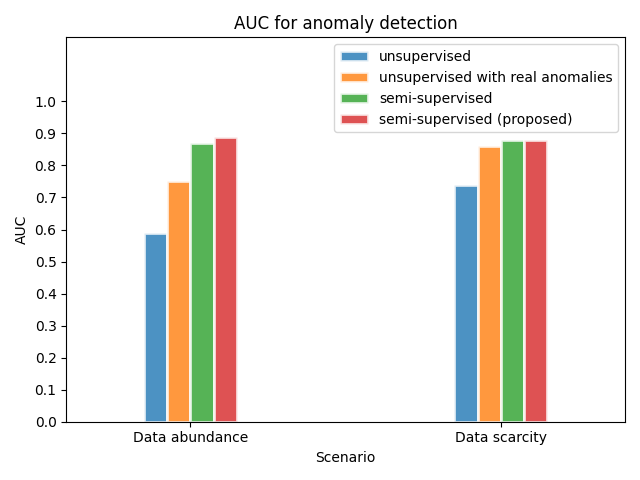}
    \caption{\rev{AUC of FCDD with unsupervised and semi-supervised training for two insulator inspection datasets (data abundance and data scarcity). The proposed semi-supervised model uses the compound modified FCDD-focal loss ($\gamma=0.4$ for data abundance $\gamma=1$ for data scarcity).}}
    \label{fig:auc_scenarios}
\end{figure}

\paragraph{Data scarcity} The FCDD model trained without any real anomalies provides an AUC of $0.74$ (Fig.~\ref{fig:auc_scenarios}) and an accuracy of $70.51\%$, which is considerably higher than the performance for IDID. This difference may partly be attributed to the larger number of images used to evaluate the models in IDID, which exposes the model to more variations in the images. \rev{In agreement with observation for IDID, an AE-based anomaly detection model outperforms FCDD with an AUC of $0.79$ and optimal accuracy of $75.46\%$.} The FCDD model trained with flashed disks as anomalies performs better with an AUC of $0.85$ and accuracy of $76.08\%$. This suggests that FCDD can leverage information about the anomalies efficiently, i.e., with only $5$ anomalies, and provide a considerable improvement in the performance of fault detection.

\subsection{\rev{Insulator inspection with semi-supervised learning}} \label{sec:results-sg}
We now present the performance of models trained with ground truth anomalies. We integrate the focal loss with modified FCDD to construct the compound loss (Eq. \ref{eq:modified_fcdd}) and train the model with different values of the focussing parameter $\gamma$. We report the results for the best performance achieved with hyperparameter tuning ($\gamma=0.4$ for data abundance and $\gamma=1$ for data scarcity) and compare with original FCDD (Eq. \ref{eq:fcdd_semi}).

\paragraph{Data abundance} \rev{The modified FCDD model trained with compound loss ($\gamma=0.4$) delivers an AUC of $0.89$ and optimal accuracy of $81.06\%$. This is an improvement of $14$ points for AUC and $10$ points for accuracy over the unsupervised model. The original FCDD model exhibits an AUC of $0.87$ and accuracy of $79\%$, performing about $2$ points worse for both metrics. This demonstrates the flexibility of modified FCDD to adapt to datasets with appropriate hyperparameters and deliver performance improvements with multiple types of anomalies. This shows that in addition to the benchmark dataset, modified FCDD can provide performance improvements by incorporating additional loss terms, a flexibility not offered by the original FCDD.}

\paragraph{Data scarcity} \rev{The modified FCDD model ($\gamma=1$), has an AUC of $0.88$ and accuracy of $81.84\%$, outperforming the unsupervised model by $3(5.84)$ points in AUC (accuracy).} This highlights the value of using ground truth anomaly maps to inform the training process about the specific patterns of faults (see Fig.~\ref{fig:auc_scenarios}). The original FCDD model delivers a comparable performance with AUC of $0.88$ and optimal accuracy of $80.61\%$.  \rev{Thus, the modified FCDD loss provides a more intuitive loss function for training and performs comparably for this dataset.}

\subsection{Improving insulator inspection in data-scarce scenario} \label{sec:results-idid-sg}
We adopt two methods of increasing the training dataset size to improve the performance in data-scarce scenarios. In the first approach, we gradually increase the number of training anomalies to cover half of the available anomalies. In the second approach, we use data from IDID to increase the dataset size for training.

\subsubsection{\rev{Increasing training anomalies from SG dataset}}
\begin{figure}
    \centering
    \includegraphics[width=0.5\textwidth]{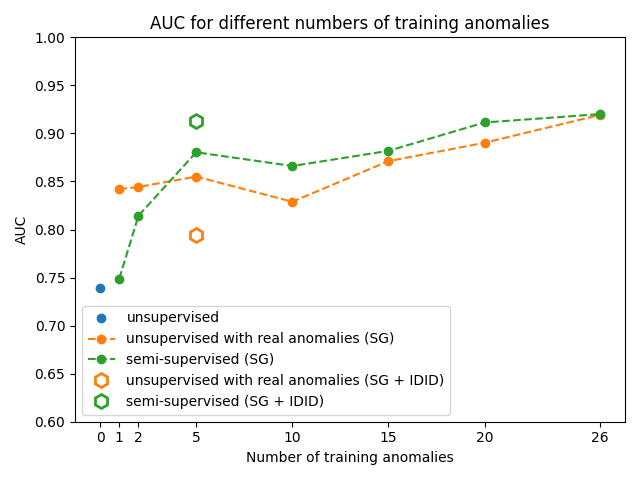}
    \caption{\rev{AUC of unsupervised and semi-supervised (modified FCDD + focal loss) models for SG dataset trained with different numbers of training anomalies.}}
    \label{fig:auc}
\end{figure}

We perform this experiment in both the unsupervised and semi-supervised (modified FCDD with focal loss) settings. The AUCs of the resulting models trained with different numbers of training anomalies are shown in Fig.~\ref{fig:auc}. The AUC of the unsupervised model without any training anomalies is also shown for reference. In the unsupervised setting, the model shows a remarkable gain of $11$ points in AUC with only one training anomaly, demonstrating the sample efficiency of the approach. We observe an increasing trend in the performance with the number of training anomalies -- except for $10$ training anomalies, which could potentially be due to the choice of particular anomalous samples for training. In the semi-supervised setting, however, we observe poor performance compared to the unsupervised setting for $1$ and $2$ training anomalies. They show an increasing trend in performance with more anomalies, outperforming the unsupervised models with $5$ or more training anomalies. This suggests that semi-supervised training, while in general better than unsupervised training, requires a minimum number of samples to learn better features. \rev{The model trained with $26$ anomalies provides an AUC of $0.92$ and an optimal accuracy of $83.37\%$ with unsupervised training. The corresponding semi-supervised model provides an AUC of $0.92$ and an accuracy of $86.49\%$.}


\subsubsection{\rev{Increasing training anomalies from IDID}} In the above approach, the number of anomalies used for testing the model is different across the scenarios. In order to maintain the same test images and increase the number of training anomalies, we leverage data from IDID. Specifically, we use all images of flashed disks in IDID, in addition to the $5$ training anomalies, and train unsupervised and semi-supervised FCDD models. The unsupervised model provides an AUC of $0.79$ with an accuracy of $72.37\%$, which is poorer than the corresponding model trained with only $5$ anomalies of the SG dataset. \rev{The semi-supervised model provides an AUC of $0.91$ and an accuracy of $84.32\%$, which is better than the corresponding model trained with only the SG dataset by $3$ points for both metrics. Note that to achieve comparable performance, $26$ images from the SG dataset are needed, which is half of the number of total anomalies in the dataset. However, by leveraging images from IDID, only $5$ anomalous images from the SG dataset were used. Thus, This observation further reinforces the value of semi-supervised training and also shows that in a data-scarce setting, the performance of the modified FCDD can be considerably improved by leveraging images from a related dataset.} 


\subsection{Explanations of FCDD}
\begin{figure*}
    \centering
    \subfigure[Explanations for IDID (left: healthy, right: faulty).]{
        \includegraphics[width=0.45\textwidth]{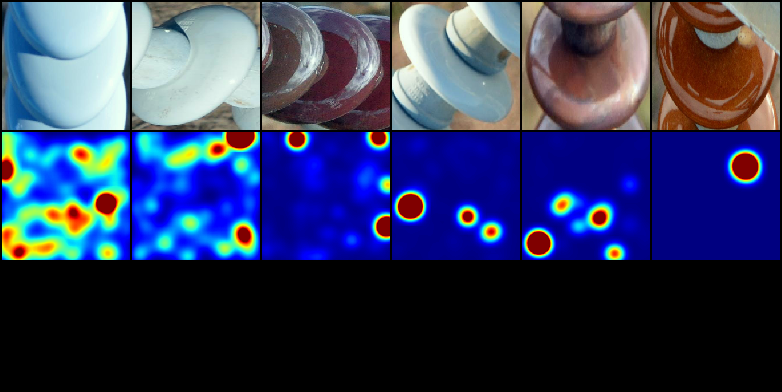}
        \includegraphics[width=0.45\textwidth]{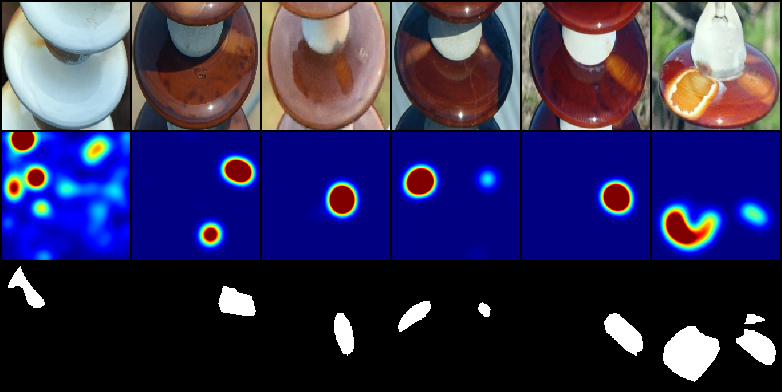}
    }
    \subfigure[Explanations for SG dataset (left: healthy, right: faulty).]{
        \includegraphics[width=0.45\textwidth]{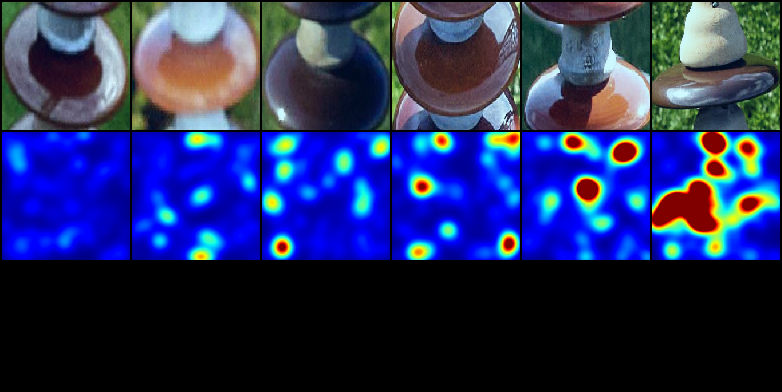}
        \includegraphics[width=0.45\textwidth]{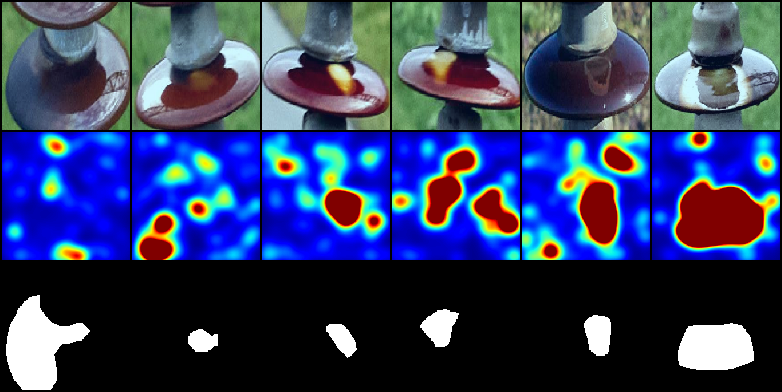}
    }
    \caption{Explanations of FCDD models for IDID (data abundant) and SG dataset (data scarce), trained in a semi-supervised manner. In each panel, the top, middle and bottom rows show the original image, explanations provided by the FCDD model, and ground truth anomaly maps, respectively. The explanations are shown as heat maps, with higher values (red) representing faulty patterns and lower values (blue) representing healthy patterns.}
    \label{fig:explanation}
\end{figure*}

The explanations from an FCDD model are obtained by up-sampling the model's output features ($z$) to match the original image size through a deterministic non-trainable Gaussian kernel-based approach \cite{liznerski2021explainable}. The up-sampled features ($\tilde{z}$) are then represented as a heat map to highlight the anomalous regions predicted by the model. Fig.~\ref{fig:explanation} shows the explanations of the semi-supervised models on sample test images for both datasets. For each dataset, Fig. \ref{fig:explanation} shows six sample test images (healthy in the left panel and faulty in the right panel), their explanations produced by the model, and the corresponding ground truth anomaly maps. It can be observed that the explanations for the faulty images match closely with the ground truth anomaly maps for the images from IDID. This suggests that the model can localise patches of discolouration on the disks with very good accuracy. The model also predicts anomalous regions on the disks that are not aligned with the ground truth anomaly maps. The explanations for the healthy images (without ground truth anomaly maps), on the other hand, consistently contain anomalous regions predicted by the model.

The explanations for the faulty images of the SG dataset are also well aligned with the ground truth maps, although they exhibit slightly more deviation from the ground truth anomaly maps compared to IDID. This can also be observed for the healthy images in this dataset. The qualitatively poor explanations for the SG dataset compared to IDID can be corroborated by the difference in AUC and optimal accuracy of the two models trained with these datasets. Although the models make false predictions for both datasets, i.e., spurious anomalous regions (on disks and in the background), these predictions are averaged to decide whether the image as a whole is normal or anomalous. This aggregation and classification based on the optimal threshold finally provides accuracies of more than $80\%$ (for the semi-supervised case). Thus, at the time of deployment, it is important to trust the predicted anomalous regions only for images classified as anomalous. However, avoiding spurious predictions altogether is an important step towards improving the anomaly detection model and remains an open challenge to employ FCDD for anomaly detection. 

\rev{
\section{Discussion} \label{sec:discussion}
In this section, we present a discussion of the key findings and highlight directions for future research.
}

\rev{
\subsection{Key findings}
We note the following key findings from the results:
\begin{enumerate}
    \item Modified FCDD improves performance over original FCDD for the MVTec-AD dataset, pushing the average performance by $0.47\%$ and worst-case performance by $2\%$.
    \item The AE model outperforms unsupervised FCDD trained without anomalies for insulator inspection. However, when trained with real anomalies, unsupervised FCDD significantly improves, outperforming AE.
    \item The semi-supervised FCDD performs better than unsupervised FCDD with as few as $5$ training anomalies. The modified FCDD performs comparably with the original FCDD. It also allows crafting composite loss functions, and such a loss function (modified FCDD with focal loss) can improve the pixel-level performance. Semi-supervised training with modified FCDD loss is thus a superior approach for insulator inspection that requires very little labelling efforts.
    \item In the data-scarce scenario, augmenting the dataset with anomalous images from a different dataset leads to a drop in performance, potentially due to domain shift.
    \item One of the biggest strengths of FCDD is the explainability of the model in terms of the predictions of faulty regions in the images. However, we observed that the explanations produced by the FCDD models, although highly accurate for several samples, can provide spurious predictions of faulty regions in some examples. Moreover, these predictions can be on the disk or in the background of the image. Thus, current explanations should be trusted more when the image has been classified as anomalous while exercising caution for those classified as normal. 
\end{enumerate}
}
\rev{
\subsection{Effect of object detection on anomaly detection model}
In this article, we have focussed our attention on the performance of the anomaly detection model with different training methods and loss functions. The proposed framework, however, consists of two stages, i.e., object detection followed by anomaly detection. A challenge with two-stage frameworks is that the errors in the first stage can propagate into the second stage, affecting the overall performance. However, as discussed in Section \ref{sec:motivation}, the localisation errors of YOLOv5 are very low, and misclassification of healthy and faulty disks contributes more to the errors. Since we eliminate the classification task from the first stage, the errors that can propagate into the second stage are only due to deviations in bounding box predictions or missed disks, both of which are very small for YOLOv5. Further, since CNNs are shift-invariant, the effect of bounding box deviations is expected to be minimal. Finally, with rapid advancement in the YOLO family of object detection models \cite{li2022yolov6,wang2023yolov7,Jocher_Ultralytics_YOLO_2023,wang2024yolov9,wang2024yolov10}, the performance of the object detection stage can be further improved, reducing the impact of errors in object detection on anomaly detection, and thus inspection of insulators.
}
\subsection{Future work}
The two-stage approach proposed in this work merges object detection with anomaly detection to identify rarely occurring insulator faults. At the time of deployment, the predictions from the object detection model can be used as input to the anomaly detection model. However, at the time of training, the parameters of the two models are learnt independently, i.e., without any shared parameters. Since the first stage aims at detecting insulators and disks, the model in the second stage (that deals with detecting patterns on disks) can potentially benefit from the learnt parameters of the object detection model. Such a study has not been conducted in this work, and can be explored in the future to potentially improve the performance and reduce the training time of the anomaly detection model.

We observed that including the faulty images from IDID did not provide a significant improvement in performance for the SG dataset. This could potentially be due to domain shift, which has not been investigated in this work. Future studies can also be directed towards incorporating domain adaptation methods in the anomaly detection stage to improve the performance in data-scarce applications.


\section{Conclusion} \label{sec:conclusions}
In this work, we proposed a two-stage approach that uses object detection and anomaly detection to reliably identify faults in insulators. Two datasets, representing data abundance and data scarcity were used in the study. A deep neural network-based explainable one-class classifier, i.e., fully convolutional data description (FCDD) was adopted to perform anomaly detection. FCDD allows both unsupervised and semi-supervised training, which enables informing the training process about the nature of faults observed in real anomalies. The FCDD model achieves AUCs of up to $0.9$ for the data-abundant application and $0.88$ for the data-scarce application. \rev{A novel loss function that is well-aligned with BCE loss was developed, which also enables integration with multiple loss terms into composite loss functions. We demonstrated that the BCE-aligned loss improves the performance of anomaly detection for the MVTec-AD dataset, pushing both the worst-case and average performance. This loss also improved the performance of insulator inspection in the data-abundant scenario. When combined with focal loss, it improved the performance of insulator inspection in the data-scarce scenario.} The semi-supervised models perform better than the unsupervised model for both datasets, highlighting the value of providing real anomalies and ground truth anomaly maps during training. However, the number of such anomalies is only $200$ for the data-abundant case and $5$ for the data-scarce case, which does not lead to a large labelling requirement for training the model. 

The proposed approach provides a method to detect rarely occurring incipient faults, specifically flashed and broken disks in both data-abundant and data-scarce applications, with minimal labelling efforts. The explanations of the model were also seen to match with the ground truth, especially for the anomalous samples. 
Improving the explanations of the model and accounting for domain shift are some open questions that can be addressed in the future.


%



\section*{Acknowledgment}
\label{Acknowledgements}
This work was supported by the Swiss Federal Office of Energy: “IMAGE - Intelligent Maintenance of Transmission Grid Assets” (Project Nr. SI/502073-01). The authors would additionally like to thank the student assistant Ms. Athina Nisioti for generating the ground truth labels for the SG dataset.

\ifCLASSOPTIONcaptionsoff
  \newpage
\fi



\bibliographystyle{IEEEtran}
\bibliography{cas-refs}

\begin{thebibliography}{10}
\providecommand{\url}[1]{#1}
\csname url@samestyle\endcsname
\providecommand{\newblock}{\relax}
\providecommand{\bibinfo}[2]{#2}
\providecommand{\BIBentrySTDinterwordspacing}{\spaceskip=0pt\relax}
\providecommand{\BIBentryALTinterwordstretchfactor}{4}
\providecommand{\BIBentryALTinterwordspacing}{\spaceskip=\fontdimen2\font plus
\BIBentryALTinterwordstretchfactor\fontdimen3\font minus \fontdimen4\font\relax}
\providecommand{\BIBforeignlanguage}[2]{{%
\expandafter\ifx\csname l@#1\endcsname\relax
\typeout{** WARNING: IEEEtran.bst: No hyphenation pattern has been}%
\typeout{** loaded for the language `#1'. Using the pattern for}%
\typeout{** the default language instead.}%
\else
\language=\csname l@#1\endcsname
\fi
#2}}
\providecommand{\BIBdecl}{\relax}
\BIBdecl

\bibitem{liu2023insulator}
Y.~Liu, D.~Liu, X.~Huang, and C.~Li, ``Insulator defect detection with deep learning: A survey,'' \emph{IET Generation, Transmission \& Distribution}, vol.~17, no.~16, pp. 3541--3558, 2023.

\bibitem{liu2018insulator}
X.~Liu, H.~Jiang, J.~Chen, J.~Chen, S.~Zhuang, and X.~Miao, ``Insulator detection in aerial images based on faster regions with convolutional neural network,'' in \emph{2018 IEEE 14th International Conference on Control and Automation (ICCA)}.\hskip 1em plus 0.5em minus 0.4em\relax IEEE, 2018, pp. 1082--1086.

\bibitem{jiang2019insulator}
H.~Jiang, X.~Qiu, J.~Chen, X.~Liu, X.~Miao, and S.~Zhuang, ``Insulator fault detection in aerial images based on ensemble learning with multi-level perception,'' \emph{IEEE Access}, vol.~7, pp. 61\,797--61\,810, 2019.

\bibitem{zhang2021insudet}
X.~Zhang, Y.~Zhang, J.~Liu, C.~Zhang, X.~Xue, H.~Zhang, and W.~Zhang, ``Insudet: A fault detection method for insulators of overhead transmission lines using convolutional neural networks,'' \emph{IEEE Transactions on Instrumentation and Measurement}, vol.~70, pp. 1--12, 2021.

\bibitem{liu2021box}
X.~Liu, X.~Miao, H.~Jiang, and J.~Chen, ``Box-point detector: A diagnosis method for insulator faults in power lines using aerial images and convolutional neural networks,'' \emph{IEEE Transactions on Power Delivery}, vol.~36, no.~6, pp. 3765--3773, 2021.

\bibitem{zhao2021insulator}
W.~Zhao, M.~Xu, X.~Cheng, and Z.~Zhao, ``An insulator in transmission lines recognition and fault detection model based on improved faster rcnn,'' \emph{IEEE Transactions on Instrumentation and Measurement}, vol.~70, pp. 1--8, 2021.

\bibitem{deng2022research}
F.~Deng, Z.~Xie, W.~Mao, B.~Li, Y.~Shan, B.~Wei, and H.~Zeng, ``Research on edge intelligent recognition method oriented to transmission line insulator fault detection,'' \emph{International Journal of Electrical Power \& Energy Systems}, vol. 139, p. 108054, 2022.

\bibitem{hao2022insulator}
K.~Hao, G.~Chen, L.~Zhao, Z.~Li, Y.~Liu, and C.~Wang, ``An insulator defect detection model in aerial images based on multiscale feature pyramid network,'' \emph{IEEE Transactions on Instrumentation and Measurement}, vol.~71, pp. 1--12, 2022.

\bibitem{shakiba2022transfer}
F.~M. Shakiba, S.~M. Azizi, and M.~Zhou, ``A transfer learning-based method to detect insulator faults of high-voltage transmission lines via aerial images: distinguishing intact and broken insulator images,'' \emph{IEEE Systems, Man, and Cybernetics Magazine}, vol.~8, no.~4, pp. 15--25, 2022.

\bibitem{dai2022uncertainty}
Z.~Dai, ``Uncertainty-aware accurate insulator fault detection based on an improved yolox model,'' \emph{Energy Reports}, vol.~8, pp. 12\,809--12\,821, 2022.

\bibitem{zhou2023fault}
M.~Zhou, B.~Li, J.~Wang, and S.~He, ``Fault detection method of glass insulator aerial image based on the improved yolov5,'' \emph{IEEE Transactions on Instrumentation and Measurement}, 2023.

\bibitem{das2022object}
L.~Das, M.~H. Saadat, B.~Gjorgiev, E.~Auger, and G.~Sansavini, ``Object detection-based inspection of power line insulators: Incipient fault detection in the low data-regime,'' 2022.

\bibitem{pang2021deep}
G.~Pang, C.~Shen, L.~Cao, and A.~V.~D. Hengel, ``Deep learning for anomaly detection: A review,'' \emph{ACM computing surveys (CSUR)}, vol.~54, no.~2, pp. 1--38, 2021.

\bibitem{hassan2022anomaly}
M.~U. Hassan, M.~H. Rehmani, and J.~Chen, ``Anomaly detection in blockchain networks: A comprehensive survey,'' \emph{IEEE Communications Surveys \& Tutorials}, 2022.

\bibitem{hilal2022financial}
W.~Hilal, S.~A. Gadsden, and J.~Yawney, ``Financial fraud: a review of anomaly detection techniques and recent advances,'' \emph{Expert systems With applications}, vol. 193, p. 116429, 2022.

\bibitem{schmidl2022anomaly}
S.~Schmidl, P.~Wenig, and T.~Papenbrock, ``Anomaly detection in time series: a comprehensive evaluation,'' \emph{Proceedings of the VLDB Endowment}, vol.~15, no.~9, pp. 1779--1797, 2022.

\bibitem{hao2021hybrid}
W.~Hao, T.~Yang, and Q.~Yang, ``Hybrid statistical-machine learning for real-time anomaly detection in industrial cyber-physical systems,'' \emph{IEEE Transactions on Automation Science and Engineering}, 2021.

\bibitem{maggipinto2022deep}
M.~Maggipinto, A.~Beghi, and G.~A. Susto, ``A deep convolutional autoencoder-based approach for anomaly detection with industrial, non-images, 2-dimensional data: A semiconductor manufacturing case study,'' \emph{IEEE Transactions on Automation Science and Engineering}, vol.~19, no.~3, pp. 1477--1490, 2022.

\bibitem{li2023self}
D.~Li, J.~Lu, T.~Zhang, and J.~Ding, ``Self-supervised learning and multisource heterogeneous information fusion based quality anomaly detection for heavy-plate shape,'' \emph{IEEE Transactions on Automation Science and Engineering}, 2023.

\bibitem{schnell2022robigan}
T.~Schnell, K.~Bott, L.~Puck, T.~Buettner, A.~Roennau, and R.~Dillmann, ``Robigan: A bidirectional wasserstein gan approach for online robot fault diagnosis via internal anomaly detection,'' in \emph{2022 IEEE/RSJ International Conference on Intelligent Robots and Systems (IROS)}.\hskip 1em plus 0.5em minus 0.4em\relax IEEE, 2022, pp. 4332--4337.

\bibitem{chen2022fault}
Y.~Chen, Z.~Zhao, H.~Wu, X.~Chen, Q.~Xiao, and Y.~Yu, ``Fault anomaly detection of synchronous machine winding based on isolation forest and impulse frequency response analysis,'' \emph{Measurement}, vol. 188, p. 110531, 2022.

\bibitem{fang2023toward}
Y.~Fang, H.~Min, X.~Wu, X.~Lei, S.~Chen, R.~Teixeira, and X.~Zhao, ``Toward interpretability in fault diagnosis for autonomous vehicles: Interpretation of sensor data anomalies,'' \emph{IEEE Sensors Journal}, 2023.

\bibitem{liznerski2021explainable}
\BIBentryALTinterwordspacing
P.~Liznerski, L.~Ruff, R.~A. Vandermeulen, B.~J. Franks, M.~Kloft, and K.~R. Muller, ``Explainable deep one-class classification,'' in \emph{International Conference on Learning Representations}, 2021. [Online]. Available: \url{https://openreview.net/forum?id=A5VV3UyIQz}
\BIBentrySTDinterwordspacing

\bibitem{ruff2021unifying}
L.~Ruff, J.~R. Kauffmann, R.~A. Vandermeulen, G.~Montavon, W.~Samek, M.~Kloft, T.~G. Dietterich, and K.-R. M{\"u}ller, ``A unifying review of deep and shallow anomaly detection,'' \emph{Proceedings of the IEEE}, vol. 109, no.~5, pp. 756--795, 2021.

\bibitem{zhou2017anomaly}
C.~Zhou and R.~C. Paffenroth, ``Anomaly detection with robust deep autoencoders,'' in \emph{Proceedings of the 23rd ACM SIGKDD international conference on knowledge discovery and data mining}, 2017, pp. 665--674.

\bibitem{zhao2017spatio}
Y.~Zhao, B.~Deng, C.~Shen, Y.~Liu, H.~Lu, and X.-S. Hua, ``Spatio-temporal autoencoder for video anomaly detection,'' in \emph{Proceedings of the 25th ACM international conference on Multimedia}, 2017, pp. 1933--1941.

\bibitem{bergmann2019mvtec}
P.~Bergmann, M.~Fauser, D.~Sattlegger, and C.~Steger, ``Mvtec ad--a comprehensive real-world dataset for unsupervised anomaly detection,'' in \emph{Proceedings of the IEEE/CVF conference on computer vision and pattern recognition}, 2019, pp. 9592--9600.

\bibitem{lin2017focal}
T.-Y. Lin, P.~Goyal, R.~Girshick, K.~He, and P.~Doll{\'a}r, ``Focal loss for dense object detection,'' in \emph{Proceedings of the IEEE international conference on computer vision}, 2017, pp. 2980--2988.

\bibitem{epriDS2021}
P.~Kulkarni, T.~Shaw, and D.~Lewis, ``Insulator defect image dataset - version 1.2: Documentation {EPRI}, {P}alo {A}lto, {CA}: 2020. 3002017949.''

\bibitem{Wada_Labelme_Image_Polygonal}
\BIBentryALTinterwordspacing
K.~Wada, ``{Labelme: Image Polygonal Annotation with Python}.'' [Online]. Available: \url{https://github.com/wkentaro/labelme}
\BIBentrySTDinterwordspacing

\bibitem{simonyan2014very}
K.~Simonyan and A.~Zisserman, ``Very deep convolutional networks for large-scale image recognition,'' \emph{arXiv preprint arXiv:1409.1556}, 2014.

\bibitem{li2022yolov6}
C.~Li, L.~Li, H.~Jiang, K.~Weng, Y.~Geng, L.~Li, Z.~Ke, Q.~Li, M.~Cheng, W.~Nie \emph{et~al.}, ``Yolov6: A single-stage object detection framework for industrial applications,'' \emph{arXiv preprint arXiv:2209.02976}, 2022.

\bibitem{wang2023yolov7}
C.-Y. Wang, A.~Bochkovskiy, and H.-Y.~M. Liao, ``Yolov7: Trainable bag-of-freebies sets new state-of-the-art for real-time object detectors,'' in \emph{Proceedings of the IEEE/CVF conference on computer vision and pattern recognition}, 2023, pp. 7464--7475.

\bibitem{Jocher_Ultralytics_YOLO_2023}
\BIBentryALTinterwordspacing
G.~Jocher, A.~Chaurasia, and J.~Qiu, ``{Ultralytics YOLO},'' Jan. 2023. [Online]. Available: \url{https://github.com/ultralytics/ultralytics}
\BIBentrySTDinterwordspacing

\bibitem{wang2024yolov9}
C.-Y. Wang, I.-H. Yeh, and H.-Y.~M. Liao, ``Yolov9: Learning what you want to learn using programmable gradient information,'' \emph{arXiv preprint arXiv:2402.13616}, 2024.

\bibitem{wang2024yolov10}
A.~Wang, H.~Chen, L.~Liu, K.~Chen, Z.~Lin, J.~Han, and G.~Ding, ``Yolov10: Real-time end-to-end object detection,'' \emph{arXiv preprint arXiv:2405.14458}, 2024.

\end{thebibliography}
\end{document}